\documentclass{article}

\PassOptionsToPackage{numbers, compress}{natbib}
\usepackage[final]{neurips_2025}

\usepackage[utf8]{inputenc} 
\usepackage[T1]{fontenc}    
\usepackage{hyperref}       
\usepackage{url}            
\usepackage{booktabs}       
\usepackage{amsfonts}       
\usepackage{nicefrac}       
\usepackage{microtype}      
\usepackage{xcolor}         
\usepackage{subcaption}
\usepackage{graphicx}
\usepackage{amsmath}
\usepackage{natbib}

\usepackage{algorithm}
\usepackage{algorithmic}

\title{Differentiation Strategies for Acoustic Inverse Problems: 
Admittance Estimation and Shape Optimization}

\author{
  Nikolas Borrel-Jensen \\
  Pasteur Labs\\
  Brooklyn, NY USA \\
  \texttt{nikolas.borrel@simulation.science} \\
  \And
  Josiah Bjorgaard \\
  Pasteur Labs\\
  Brooklyn, NY USA \\
  \texttt{josiah.bjorgaard@simulation.science} \\
}

\begin{document}
\maketitle

\begin{abstract}
We demonstrate a practical differentiable programming approach for acoustic inverse problems through two applications: admittance estimation and shape optimization for resonance damping. First, we show that JAX-FEM's automatic differentiation (AD) enables direct gradient-based estimation of complex boundary admittance from sparse pressure measurements, achieving 3-digit precision without requiring manual derivation of adjoint equations. Second, we apply randomized finite differences to acoustic shape optimization, combining JAX-FEM for forward simulation with PyTorch3D for mesh manipulation through AD. By separating physics-driven boundary optimization from geometry-driven interior mesh adaptation, we achieve 48.1\% energy reduction at target frequencies with 30-fold fewer FEM solutions compared to standard finite difference on the full mesh. This work showcases how modern differentiable software stacks enable rapid prototyping of optimization workflows for physics-based inverse problems, with automatic differentiation for parameter estimation and a combination of finite differences and AD for geometric design.
\end{abstract}

\section{Introduction}

Differentiable programming has revolutionized machine learning, but its potential for scientific computing remains underexplored \cite{baydin2018automatic, Lavin2021SimulationIT}. While automatic differentiation (AD) through neural networks is standard practice, applying AD to traditional physics simulators (finite elements, finite differences, etc.) introduces unique challenges: sparse linear solvers, complex geometries, and expensive forward passes \cite{holl2020diffpd}. This paper demonstrates how modern differentiable software stacks enable efficient solutions to acoustic inverse problems through appropriate choice of differentiation strategies.

\textbf{Problem 1: Admittance Estimation.} Acoustic boundary conditions are characterized by complex admittance values $\beta = \beta_r + i\beta_i$ representing material absorption and reactivity \cite{kuttruff2016room}. Estimating these parameters from sparse pressure measurements requires solving an inverse problem: given measurements $\{p_i^{\text{meas}}\}$, find $\beta$ such that the forward model predictions match observations. Traditional approaches require deriving and implementing adjoint equations for gradient computation. Leveraging JAX-FEM's (\cite{xue2023jaxfem}) automatic differentiation to compute exact gradients $\nabla_\beta J$ we eliminate manual adjoint derivation and efficiently compute gradients for parameter optimization.

\textbf{Problem 2: Shape Optimization.} Room acoustics are strongly affected by geometry \cite{kuttruff2016room}. Square rooms exhibit resonances at specific frequencies, creating hot spots that degrade sound quality. Optimizing room shape to minimize energy at undesired frequencies requires repeatedly solving the Helmholtz equation. We efficiently enable a hybrid strategy where randomized finite differences for boundary optimization are combined with automatic differentiation for interior mesh morphing.

This paper demonstrates how to use differentiable physics (JAX-FEM \cite{xue2023jaxfem}) and differentiable geometry (PyTorch3D \cite{ravi2020pytorch3d}) to enable practical optimization workflows in acoustics. Our results show: (1) 3-digit precision admittance estimation from 25 sparse measurements using AD, and (2) 48.1\% resonance energy reduction with 30-fold speedup over standard finite difference (FD) optimization.

\section{Theory}
\subsection{Acoustic Forward Model}

The acoustic pressure field $p(\mathbf{x})$ in a 2D domain $\Omega$ with frequency $f$ is governed by the Helmholtz equation \cite{kuttruff2016room}

\begin{equation}
\nabla^2 p + k^2 p = -f(\mathbf{x}) \quad \text{in } \Omega,
\end{equation}

where $k = 2\pi f/c$ is the wavenumber, $c$ the speed of sound, and $f(\mathbf{x})$ a source term. We model a Gaussian point source at location $\mathbf{s}$ as

\begin{equation}
f(\mathbf{x}) = A \exp\left(-\frac{|\mathbf{x} - \mathbf{s}|^2}{2\sigma^2}\right). \label{eq:gaussian_source}
\end{equation}

At the boundaries $\partial\Omega$, we impose an admittance boundary condition $\frac{\partial p}{\partial n} + ik\beta p = 0 \text{ on } \partial\Omega$, where $\beta = \beta_r + i\beta_i$ is the complex normalized admittance (dimensionless) representing wall absorption. The real part $\beta_r$ characterizes energy dissipation, while the imaginary part $\beta_i$ accounts for reactive effects.

\subsection{Inverse Problem Formulation}

Given sparse pressure measurements $\{p_i^{\text{meas}}\}$ at locations $\{\mathbf{x}_i\}$, we seek to estimate $\beta$ by minimizing the weighted least-squares objective:

\begin{equation}
J(\beta) = w_{\text{mag}} \sum_i \left(|p_i^{\text{pred}}| - |p_i^{\text{meas}}|\right)^2 + w_{\text{phase}} \sum_i \left(\angle p_i^{\text{pred}} - \angle p_i^{\text{meas}}\right)^2
+ w_{\text{rel}} \left(\frac{\|\mathbf{p}^{\text{pred}} - \mathbf{p}^{\text{meas}}\|_2}{\|\mathbf{p}^{\text{meas}}\|_2}\right)^2, \label{eq:admittance_loss}
\end{equation}

where $p_i^{\text{pred}}$ denotes the predicted pressure from the forward model with current $\beta$ estimate.

\subsection{Gradient-Based Optimization with Autodifferentiation}

We employ gradient-based optimization to minimize $J(\beta)$, requiring computation of $\nabla_\beta J$. Traditional approaches necessitate deriving and implementing adjoint equations \cite{baydin2018automatic}. Instead, we leverage the autodifferentiation capabilities of JAX-FEM, which automatically computes exact gradients through the entire FEM solve via reverse-mode differentiation. This transforms the complete forward solve $\beta \to \mathbf{K}(\beta) \to \mathbf{p}^{\text{pred}} \to J$ into a differentiable computational graph, enabling efficient gradient computation without manual derivation of sensitivity equations. The matrix $\mathbf{K}$ is a complex matrix explained in Appendix \ref{sec:finite_element_discretization}.

\subsection{Geometry optimization}
We define the PDE-constrained shape optimization problem as
\begin{equation}
\min_{\partial\Omega} J(\partial\Omega); \mathcal{F}(u; \Omega) = 0, \label{eq:shape_opt}
\end{equation}
where $\mathcal{F}$ is the governing PDE constraint and $u$ is the state variable . JAX-FEM does not implement AD w.r.t. the mesh. To perform optimization with \autoref{eq:shape_opt} we instead exploit a key structural property, the objective function is defined primarily through the boundary vertices $\mathbf{V}_b$ and the internal vertices $\mathbf{V}_i$ are used only for discretization. Discretizing $\Omega$ with finite elements requires $N_b$ boundary vertices and $N_i \gg N_b$ interior vertices. This structure appears across acoustics, electromagnetics, heat transfer, structural mechanics, and fluid dynamics where boundary conditions dominate performance. Instead of computing gradients w.r.t. all $N = N_b + N_i$ vertices, we can perform optimization with a two-stage hybrid framework, detailed in Appendix \ref{sec:geometry_optimization_algorithm}.

In stage 1 we perform boundary optimization with randomized gradients. To compute $\nabla_{\mathbf{V}_b} J$ efficiently without deriving adjoints, we use randomized finite differences \cite{scheinberg2022finite} based on the Johnson-Lindenstrauss lemma \cite{johnson1984extensions}. Sample $S$ random directions $\mathbf{d}_s \sim \mathcal{N}(0, I_{N_b \times d})$, compute directional derivatives $g_s = [J(\mathbf{V}_b + \epsilon \mathbf{d}_s) - J(\mathbf{V}_b)]/\epsilon$, and accumulate the gradient estimate $\hat{\nabla}_{\mathbf{V}_b} J = \frac{1}{S} \sum_{s=1}^S g_s \mathbf{d}_s$. This estimator is unbiased with variance $\propto 1/S$ and requires only $S$ PDE solves (typically $S = 20$-$50$) versus $2N_b$ for coordinate-wise finite differences. The PDE solver is treated as a black box so there is no need to implement AD or derive adjoints.

In stage 2 we perform interior optimization via AD. Given fixed boundary $\mathbf{V}_b$, we optimize $\mathbf{V}_i$ for mesh quality using 
\begin{equation}
\mathcal{L}_{\text{mesh}}(\mathbf{V}_i; \mathbf{V}_b) = w_e \mathcal{L}_{\text{edge}} + w_l \mathcal{L}_{\text{Laplacian}} + w_n \mathcal{L}_{\text{normal}} + w_A\mathcal{L}_{\text{area}}, \label{eq:loss_interior}
\end{equation} where the losses $\mathcal{L}_{\text{edge}}$ penalize edge length irregularity, $\mathcal{L}_{\text{Laplacian}}$ Laplacian non-uniformity, $\mathcal{L}_{\text{normal}}$ face normal inconsistency, and $\mathcal{L}_{\text{area}}$ is preserving the floor area. These purely geometric losses evaluate in milliseconds via autodiff and are implemented in PyTorch3D \cite{ravi2020pytorch3d}, requiring no PDE solves. Interior gradients are computed via backpropagation, boundary gradients are zeroed to keep $\mathbf{V}_b$ fixed, and standard optimizers (Adam) update interior vertices. Each iteration alternates: (1) update $\mathbf{V}_b$ via $S$ randomized gradient samples through the PDE solver, and (2) update $\mathbf{V}_i$ via $M_{\text{inner}}$ autodiff-based mesh quality steps. Total cost is $\sim$$S$ PDE solves per iteration versus $\sim$$2N$ for full-mesh coordinate-wise FD.

\section{Results}

\subsection{Admittance estimation}
We estimate the complex boundary admittance from sparse pressure measurements in a $2\text{ m} \times 2\text{ m}$ domain at frequency $f=1000$ Hz. The reference pressure field is generated with true admittance $\beta_{\text{true}} = 1.5 + 0.3i$, and $2\%$ normally distributed noise is added to simulate measurement uncertainty. A Gaussian source (defined in \autoref{eq:gaussian_source}) is placed at the domain center $\mathbf{s} = (1.0\text{ m}, 1.0\text{ m})$ with width $\sigma=0.086$ m and amplitude $A = 1000$. Measurements are collected at 3 cluster locations centered at $\mathbf{x}_0 = (0.7\text{ m}, 1.7\text{ m})$, $\mathbf{x}_1 = (1.0\text{ m}, 1.7\text{ m})$, and $\mathbf{x}_2 = (1.3\text{ m}, 1.7\text{ m})$. Each cluster includes all finite element quadrature points within a 0.1 m radius, totaling 25 measurement locations. This sparse arrangement mimics a realistic microphone array where physical size constraints limit sensor density.

We used the SGD optimizer with learning rate 0.1 for 300 iterations, starting from an initial normalized admittance of $\beta_{\text{init}} = 3.0 + 0.05i$. The weights for the loss terms in \autoref{eq:admittance_loss} are set to $w_{\text{mag}} = 0.5$, $w_{\text{phase}} = 0.1$, and $w_{\text{rel}} = 5.0$. We found it beneficial to set the weight for the relative error term large to learn the correct amplitude scaling. The optimizer converged after 300 iterations, estimating the admittance to three-digit precision at $\beta_{\text{est}} = 1.496+0.299i$, compared to the true value of $\beta_{\text{true}} = 1.5+0.3i$. This demonstrates that JAX-FEM's autodifferentiation successfully recovered the boundary admittance.

\begin{figure}
\centering
\begin{subfigure}{0.3\textwidth}
    \includegraphics[width=\textwidth]{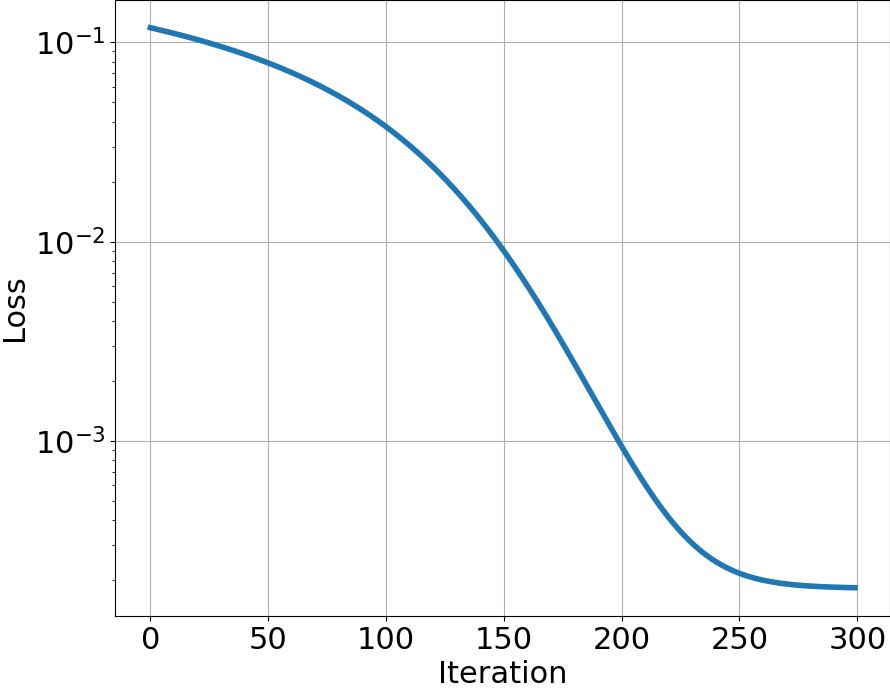}
    \caption{Loss for 300 iterations.\newline}
    \label{fig:admittance_loss}
\end{subfigure}
\hfill
\begin{subfigure}{0.3\textwidth}
    \includegraphics[width=\textwidth]{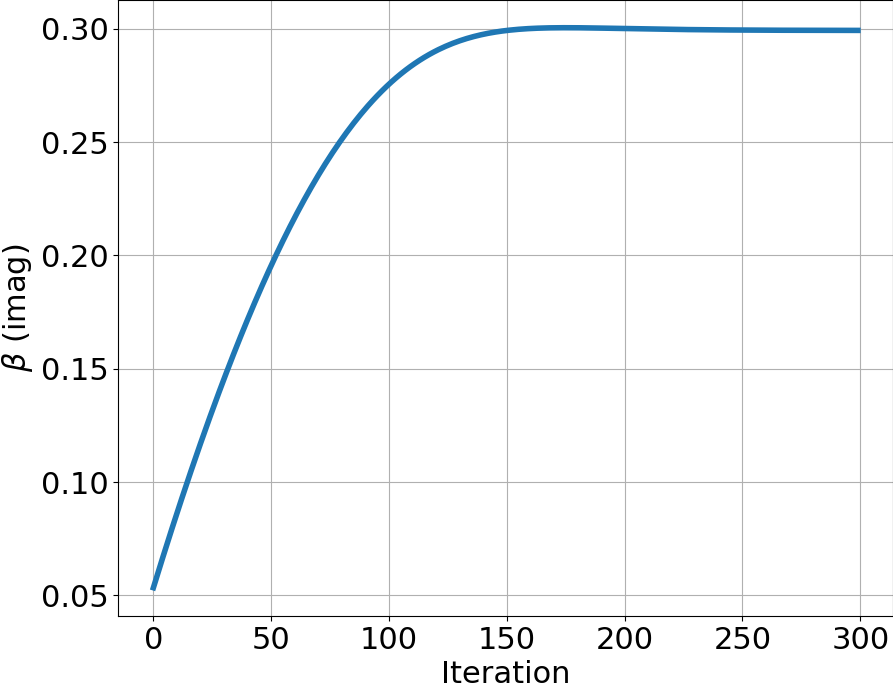}
    \caption{Imaginary part of the estimated normalized admittance $\beta$.}
    \label{fig:admittance_imag}
\end{subfigure}
\hfill
\begin{subfigure}{0.3\textwidth}
    \includegraphics[width=\textwidth]{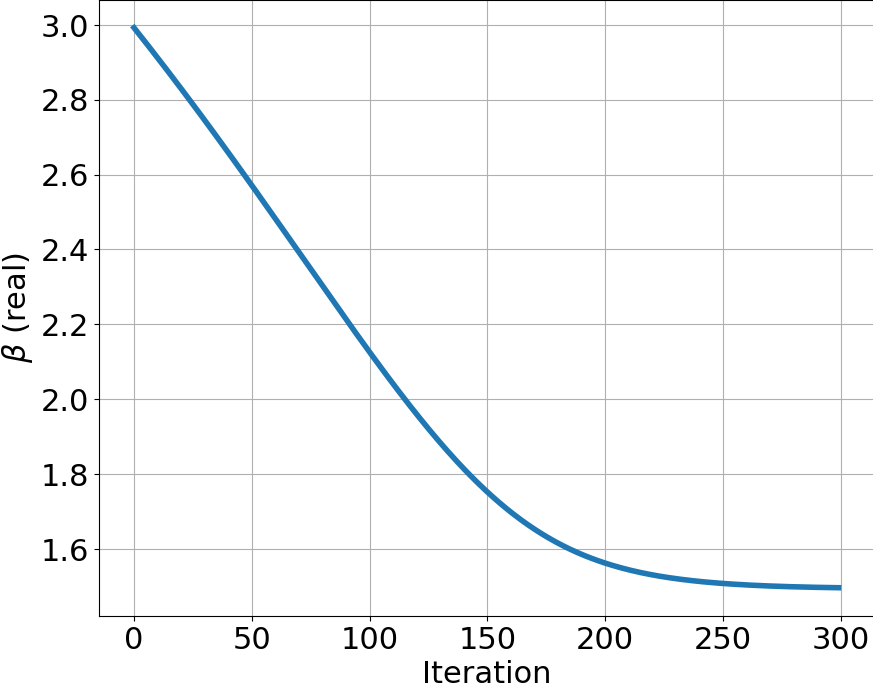}
    \caption{Real part of the estimated normalized admittance $\beta$.}
    \label{fig:admittance_real}
\end{subfigure}
\caption{Convergence of the normalized admittance estimation. The initial guess was $\beta_{\text{est}} = 3.0 + 0.05$ and the method estimated the real admittance $\beta_{\text{true}}$ within 3 digits.}
\end{figure}

\subsection{Room geometry optimization}
We demonstrate geometry optimization on room acoustics by minimizing resonances in enclosed spaces. Square/rectangular rooms exhibit strong standing waves at specific frequencies creating hot spots which are problematic for recording studios and concert halls. The total acoustic energy $\int_{\Omega} |p|^2 d\mathbf{x}$ is minimized while preserving floor area $\mathcal{L}_{\text{area}} = (|\Omega| - |\Omega_0|)^2$ in \autoref{eq:loss_interior} with $w_A=100$. We set up a test case for a 4m $\times$ 4m room, discretized into 258 vertices (68 boundary, 190 interior) and 458 triangular elements. The acoustic source is positioned at 0.5m $\times$ 0.5m from the center of the room with an amplitude factor of 1000 and frequency of 100 Hz close to the (2,1) mode. The room walls are given an absorptive boundary condition with admittance $\beta=1.5 + 0.2j$. We use separate Adam optimizers for each stage with learning rates of 0.1 and perform 20 mesh optimization steps for each boundary optimization step and 30 JAX-FEM solutions for the random gradient calculation. We note that this would require 916 solutions to determine an FD gradient, which is over a 30-fold reduction in number of JAX-FEM solutions required.

\begin{figure}
\centering
\begin{subfigure}{0.3\textwidth}
    \includegraphics[width=\textwidth]{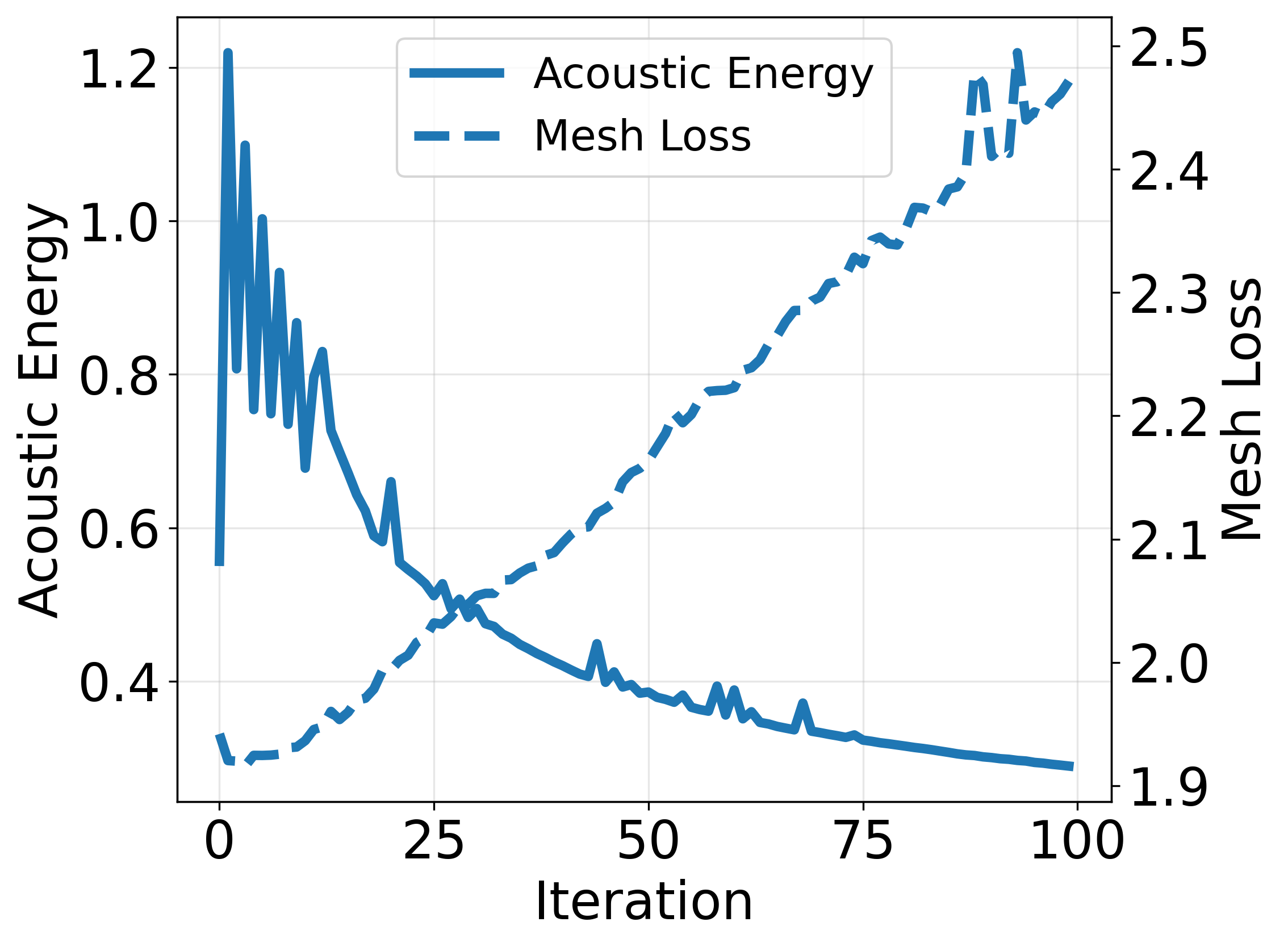}    
    \caption{Optimization metrics.} \label{fig:losses}
\end{subfigure}
\begin{subfigure}{0.3\textwidth}
    \includegraphics[width=\textwidth]{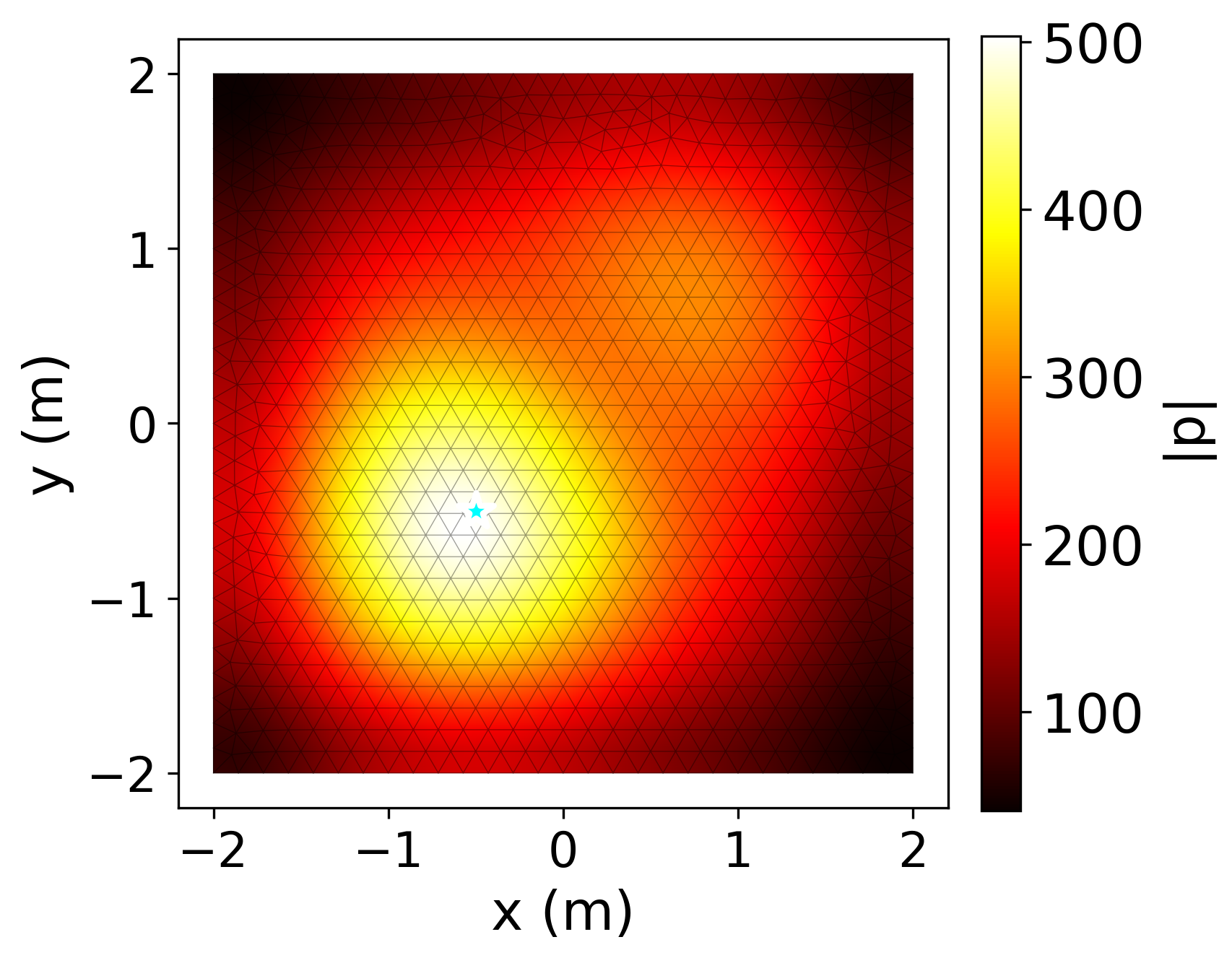}    
    \caption{Mesh and $|p|$ at step 0.} \label{fig:pressure_initial}
\end{subfigure}
\begin{subfigure}{0.3\textwidth}
    \includegraphics[width=\textwidth]{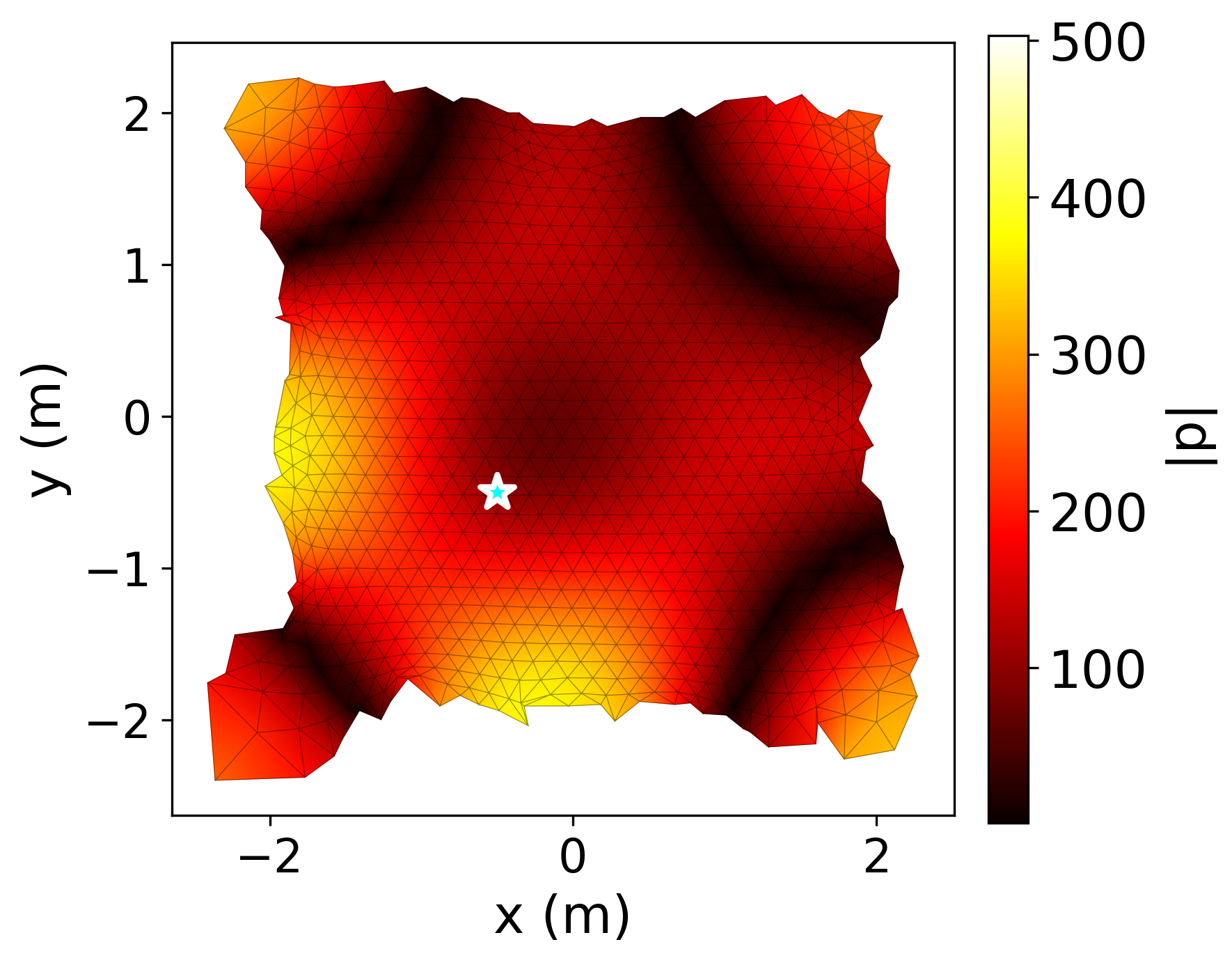}    
    \caption{Mesh and $|p|$ after at step 100.} \label{fig:pressure_final}
\end{subfigure}
\caption{Results of room geometry optimization w.r.t. acoustic energy over 100 steps, showing total acoustic energy (scaled by $10^{-8}$) and mesh loss (scaled by $10^2$). Initial mesh and $|p|$ field and mesh geometry are shown along with mesh and $|p|$ at step 100.}
\end{figure}

As shown in \autoref{fig:losses}, the acoustic energy decreases while the original mesh shown in \autoref{fig:pressure_initial} deforms in a manner that suppresses standing waves as shown in \autoref{fig:pressure_final}. The $\mathcal{L}_{\text{mesh}}$ increases by $\sim$25\% over the total optimization, which is likely due to the more complex geometry. We also note that some exploration is observed which initially increases the acoustic energy. This could be contributed to by the stochasticity in calculation of $\nabla_{\mathbf{V}_b} J$.

Avenues of further work include multiband objectives, improved mesh conditioning, and experimentation with 3 dimensional systems. Although a multiband objective targeting the lower modes would have been of particular interest, the JAX-FEM solver failed to converge for certain configurations, which limited our ability to perform multi-band optimizations. Improvements to the solver should enable this. The optimized boundary shape shown in \autoref{fig:pressure_final} lacks typical regularity (smoothness) required for real-world applications, so additional cost functions and smoothing for boundary optimization will be explored. Finally, we intend to apply the described methods to 3D environments.
\section{Conclusion}
We demonstrated how automatic differentiation and randomized FD address inverse problems in room acoustics. First, we exploited AD through a numerical FEM solver to estimate complex boundary admittance from sparse measurements, achieving 3-digit accuracy. Second, we suppressed resonant frequencies through shape optimization by combining randomized FD for boundary design with automatic differentiation for interior mesh quality. Our results showcase how selecting differentiation strategies appropriate to problem structure enables efficient solutions to physics-based inverse problems.

\newpage
\bibliographystyle{unsrt}
\bibliography{bibtex}

\newpage
\appendix

\section{Technical Appendices and Supplementary Material}
\subsection{Finite Element Discretization}\label{sec:finite_element_discretization}

Using the finite element method with basis functions $\phi_j$, we approximate $p \approx \sum_j p_j \phi_j$. The weak formulation, obtained by multiplying by test functions $v$ and integrating by parts, yields

\begin{equation}
\int_\Omega \nabla p \cdot \nabla v \, dA - k^2 \int_\Omega p v \, dA + ik\beta \int_{\partial\Omega} p v \, ds = \int_\Omega s v \, dA
\end{equation}

This results in a complex linear system $\mathbf{K}(\beta)\mathbf{p} = \mathbf{f}$, where the system matrix $\mathbf{K}$ depends on the admittance parameter $\beta$.

\subsection{ Two-Stage Randomized Gradient Algorithm for Shape Optimization }\label{sec:geometry_optimization_algorithm}

\begin{table}[H]
\centering
\begin{tabular}{ll}
\toprule
Symbol & Description \\
\midrule
$\mathbf{V}_b \in \mathbb{R}^{N_b \times d}$ & Boundary vertices \\
$\mathbf{V}_i \in \mathbb{R}^{N_i \times d}$ & Interior vertices \\
$N_b$ & Number of boundary vertices \\
$N_i$ & Number of interior vertices \\
$d$ & Spatial dimension ($d = 2$ or $3$) \\
$J(\mathbf{V}_b)$ & Objective function (PDE-constrained) \\
$S$ & Number of random samples \\
$\epsilon$ & Finite difference perturbation size \\
$\eta_b, \eta_i$ & Learning rates for boundary/interior \\
$w_e, w_l, w_n$ & Mesh quality loss weights \\
$M_{\text{inner}}$ & Interior optimization iterations per outer step \\
$K$ & Maximum outer iterations \\
$\tau$ & Convergence tolerance \\
$\mathcal{L}_{\text{edge}}$ & Edge length regularity loss \\
$\mathcal{L}_{\text{Laplacian}}$ & Laplacian smoothness loss \\
$\mathcal{L}_{\text{normal}}$ & Face normal consistency loss \\
\bottomrule
\end{tabular}
\end{table}

\begin{algorithm}[htbp]
\caption{Randomized Boundary Gradient Estimation}
\label{alg:boundary_gradient}
\begin{algorithmic}[1]
\REQUIRE $\mathbf{V}_b \in \mathbb{R}^{N_b \times d}$, $\epsilon > 0$
\ENSURE $\hat{\nabla}_{\mathbf{V}_b} J \in \mathbb{R}^{N_b \times d}$
\STATE $\hat{\nabla}_{\mathbf{V}_b} J \leftarrow \mathbf{0}$
\STATE $J_0 \leftarrow J(\mathbf{V}_b)$
\FOR{$s = 1, \ldots, S$}
    \STATE $\mathbf{d}_s \sim \mathcal{N}(\mathbf{0}, I_{N_b \times d})$
    \STATE $g_s \leftarrow \frac{J(\mathbf{V}_b + \epsilon \mathbf{d}_s) - J_0}{\epsilon}$
    \STATE $\hat{\nabla}_{\mathbf{V}_b} J \leftarrow \hat{\nabla}_{\mathbf{V}_b} J + g_s \mathbf{d}_s$
\ENDFOR
\STATE $\hat{\nabla}_{\mathbf{V}_b} J \leftarrow \frac{1}{S} \hat{\nabla}_{\mathbf{V}_b} J$
\RETURN $\hat{\nabla}_{\mathbf{V}_b} J$
\end{algorithmic}
\end{algorithm}

\begin{algorithm}[htbp]
\caption{Interior Mesh Optimization}
\label{alg:interior_opt}
\begin{algorithmic}[1]
\REQUIRE $\mathbf{V}_i \in \mathbb{R}^{N_i \times d}$, $\mathbf{V}_b \in \mathbb{R}^{N_b \times d}$, $\eta_i > 0$
\ENSURE $\mathbf{V}_i^* \in \mathbb{R}^{N_i \times d}$
\FOR{$m = 1, \ldots, M_{\text{inner}}$}
    \STATE $\mathcal{L}_{\text{mesh}}(\mathbf{V}_i; \mathbf{V}_b) \leftarrow w_e \mathcal{L}_{\text{edge}} + w_l \mathcal{L}_{\text{Laplacian}} + w_n \mathcal{L}_{\text{normal}}$
    \STATE $\mathbf{g}_i \leftarrow \nabla_{\mathbf{V}_i} \mathcal{L}_{\text{mesh}}(\mathbf{V}_i; \mathbf{V}_b)$
    \STATE $\mathbf{V}_i \leftarrow \mathbf{V}_i - \eta_i \mathbf{g}_i$
\ENDFOR
\RETURN $\mathbf{V}_i$
\end{algorithmic}
\end{algorithm}

\begin{algorithm}[htbp]
\caption{Two-Stage Hybrid Geometry Optimization}
\label{alg:geometry_opt}
\begin{algorithmic}[1]
\REQUIRE $\mathbf{V}_b^{(0)} \in \mathbb{R}^{N_b \times d}$, $\mathbf{V}_i^{(0)} \in \mathbb{R}^{N_i \times d}$,  $\tau > 0$
\ENSURE $(\mathbf{V}_b^*, \mathbf{V}_i^*)$
\STATE $k \leftarrow 0$, $J_{\text{prev}} \leftarrow \infty$
\WHILE{$k < K$ \textbf{and} $|J(\mathbf{V}_b^{(k)}) - J_{\text{prev}}| > \tau$}
    \STATE $J_{\text{prev}} \leftarrow J(\mathbf{V}_b^{(k)})$
    \STATE $\hat{\nabla}_{\mathbf{V}_b} J \leftarrow$ \textsc{BoundaryGradient}$(\mathbf{V}_b^{(k)}, S, \epsilon)$
    \STATE $\mathbf{V}_b^{(k+1)} \leftarrow \mathbf{V}_b^{(k)} - \eta_b \hat{\nabla}_{\mathbf{V}_b} J$
    \STATE $\mathbf{V}_i^{(k+1)} \leftarrow$ \textsc{InteriorOptimization}$(\mathbf{V}_i^{(k)}, \mathbf{V}_b^{(k+1)}, w_e, w_l, w_n, \eta_i, M_{\text{inner}})$
    \STATE $k \leftarrow k + 1$
\ENDWHILE
\RETURN $(\mathbf{V}_b^{(k)}, \mathbf{V}_i^{(k)})$
\end{algorithmic}
\end{algorithm}
\newpage

\end{document}